%% file: coling2018.tex
\documentclass[11pt]{article}
\usepackage{coling2018}
\usepackage{times}
\usepackage{url}
\usepackage{latexsym}

\usepackage{framed}
\usepackage{amsfonts}

\usepackage{url}
\usepackage{covington} 
\usepackage{microtype} 
\usepackage{amsmath}
\usepackage{amssymb}
\usepackage{graphicx}
\usepackage{pbox}
\usepackage{color}
\usepackage{mathtools}

\newcommand{\predicate}[1]{\texttt{#1}}

\usepackage[linesnumbered,ruled,vlined]{algorithm2e}

\title{Graph-Based Decoding for Event Sequencing and Coreference Resolution}

 \author{Zhengzhong Liu \and Teruko Mitamura \and Eduard Hovy \\
Language Technologies Institute \\
Carnegie Mellon University \\
Pittsburgh, PA 15213, USA \\
{\tt \{liu, teruko, hovy\}@cs.cmu.edu}}

\date{}

\begin{document}
\maketitle

\input{0abstract}
\input{1introduction}

\input{2related}
\input{3model}
\input{4experiment}
\input{5discussion}
\input{6conclusion}

\section*{Acknowledgements}
This research was supported in part by DARPA grant FA8750-18-2-0018 funded under the AIDA program. 


\bibliographystyle{acl}
\bibliography{acl2017}

\end{document}

%% file: 0abstract.tex
\begin{abstract}
	Events in text documents are interrelated in complex ways. In this paper, we study two types of relation: \textbf{Event Coreference} and \textbf{Event Sequencing}.
	We show that the popular tree-like decoding structure for automated Event Coreference is not suitable for Event Sequencing. To this end, we propose a graph-based decoding algorithm that is applicable to both tasks.
	The new decoding algorithm supports flexible feature sets for both tasks.
	Empirically, our event coreference system has achieved state-of-the-art performance on the TAC-KBP 2015 event coreference task and our event sequencing system beats a strong temporal-based, oracle-informed baseline. 
	We discuss the challenges of studying these event relations. 
\end{abstract}

%% file: 1introduction.tex
\section{Introduction}

%
%
\blfootnote{
    %
    %
    %
    %
	%
    \hspace{-0.65cm}  
    This work is licensed under a Creative Commons 
    Attribution 4.0 International License.
    License details:
    \url{http://creativecommons.org/licenses/by/4.0/}
}

Events are important building blocks of documents.  They play a key role in document understanding tasks, such as information extraction~\cite{Chambers2011a}, news summarization~\cite{Vossen2015}, story understanding~\cite{Mostafazadeh2016}. Conceptually, \textbf{events} correspond to state changes and 
normally include a location, a time interval, and several entities/participants. In a text, events are realized as text spans, normally as verbs and nouns that indicate state changes~\cite{Vendler1957}. The text spans are often referred to as \textbf{event mentions} or \textbf{event nuggets} (We use the term event mention in this paper.).
The textual mentions of events have rich relations among them, and collectively convey the meaning of one or more related documents. In this paper, we study two different types of relation: \textbf{Event Hopper Coreference (EH)} and \textbf{Event Sequencing (ES)}.

\textbf{Event Coreference:} There is a rich literature on the Event Coreference problem~\cite{Liu2013,Cybulska2014,Lu2016,Peng2016Detection,Lu2017,Araki2018}. By analogy to entity coreference, the ``same'' conceptual event may be realized by multiple text spans (event mentions). The coreference problem aims at identifying these relations to recover events from the text spans. The \textbf{Event Hopper} Coreference task in the TAC-KBP evaluation campaign defines coreference links as follows~\cite{Mitamura2018}: Two event mentions are considered coreferent if they refer to the conceptually same underlying event, even if their arguments are not strictly identical. For example, mentions that share similar temporal and location scope, though not necessarily the same expression, are considered to be coreferent (\textit{Attack in Baghdad on Thursday} vs. \textit{Bombing in the Green Zone last week}). This means that the event arguments of coreferential events mentions can be non-coreferential (18 killed vs. dozens killed), as long as they refer to the same event, judging from the available evidence.

\textbf{Event Sequencing:}  The coreference relations build up events from scattered mentions. On the basis of events, various other types of relations can then be established between them. The Event Sequencing task studies one such relation. The task is motivated by Schank's {\em scripts}~\cite{Schank1977}, which suggests that human organize information through procedural data structures, reassembling sequences of events. For example, the list of verbs \textit{order, eat, pay, leave} may trigger the restaurant script. A human can conduct reasoning with a typical ordering of these events based on common sense (e.g., \textit{order} should be the first event, \textit{leave} should be the last event).

\begin{figure*}
\centering
\includegraphics[width=0.85\textwidth]{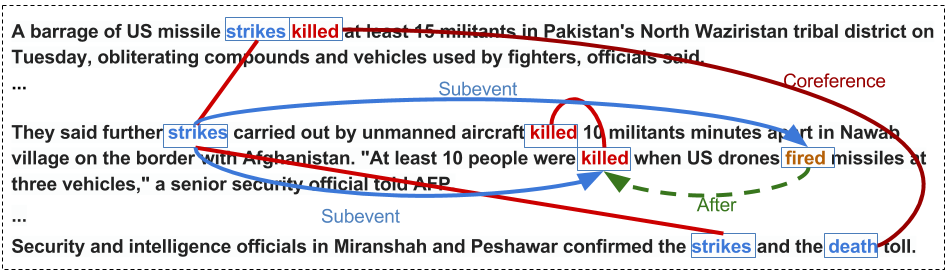}
\caption{\label{fig:example} Example of Event Coreference and Sequence relations. Red lines are coreference links; solid blue arrows represent Subevent relations; dotted green arrows represent After relations.}
\end{figure*}

The ES task studies how to group and order events from text documents belonging to the same script. Figure~\ref{fig:example} shows some annotation examples. Conceptually, event sequencing relations hold between the events, while coreference relations hold between textual event mentions.
Given a document, the ES task requires systems to identify events within the same script and classify their inter-relations. These relations can be represented as labeled Directed Acyclic Graphs (DAGs). There are two types of relations\footnote{Detailed definition of relations can be found in http://cairo.lti.cs.cmu.edu/kbp/2016/after/}: \textbf{After} relations connect events following script orders (e.g. \textit{order} followed by \textit{eating}); \textbf{Subevent} relations connect events to a larger event that contains them. 
In this paper, we focus only on the \textbf{After} relations. 

Since script-based understanding is built in the ES task, it has some unique properties comparing to pure temporal ordering: 1) event sequences from different scripts provide separate logical divisions of text, while temporal ordering considers all events to lie on a single timeline; 2) temporal relations for events occurring at similar time points may be complicated. Script-based relations may alleviate the problem. For example, if a \predicate{bombing}  \predicate{kill}s some people, the temporal relation of the \predicate{bombing} and \predicate{kill} may be ``inclusion'' or ``after''. This is considered an \textbf{After} relation in ES because \predicate{bombing} causes the \predicate{kill}ing.

For structure prediction, decoding --- recovering the complex structure from local decisions --- is one of the core problems.
The most successful decoding algorithm for coreference nowadays is mention ranking based~\cite{Bjorkelund2014,Durrett2014,Lee2017}. These models rank the antecedents (mentions that appear earlier in discourse) and recover the full coreference clusters from local decisions. However, unlike coreference relations, sequencing relations are directed. Coreference decoding algorithms cannot be directly applied to such relations (\S\ref{sec:graph_model}). To solve this problem, we propose a unified graph-based framework that tackles both event coreference and event sequencing. Our method achieves state-of-the-art results on the event coreference task (\S\ref{sec:coref_results}) and beats an informed baseline on the event sequencing task (\S\ref{sec:seq_results}). Finally, we analyze the results and discuss the difficult challenges for both tasks (\S\ref{sec:discussion}). Detailed definitions of these tasks can be found in the corresponding task documents\footnote{\url{http://cairo.lti.cs.cmu.edu/kbp/2017/event/documents}}.

%% file: 2related.tex
\section{Related Work}
Many researchers have worked on event coreference tasks since \mbox{\newcite{Humphreys1997a}}. Recent advances in event coreference have been promoted by the availability of annotated corpora. However, due to the complex nature of events, approaches to event coreference adopt quite different assumptions and definitions. Most of event coreference researches are conducted on the popular ACE corpus \cite{Chen2009,Chen2009a,Sangeetha2012,Chen2013,Chen2015}. Unlike the TAC KBP setting, the definition of event coreference in the ACE corpus requires strict argument matching. Work on the Intelligence Community (IC) Corpus \cite{Hovy2013,Cybulska2012,Liu2013,Araki2013} considers event relations on a restricted domain (i.e., terrorist events). Works on the ECB corpus \cite{Lee2012,Cybulska2014} focuses on both within-document and cross-document coreference.

Our work follows the line of work promoted by the TAC-KBP event nugget tasks \cite{Mitamura2016}. There is a small but growing amount of work on conducting event coreference on the TAC-KBP datasets \cite{Lu2016,Peng2016Detection,Lu2017}. The TAC dataset uses a relaxed coreference definition comparing to other corpora, requiring two event mentions to intuitively refer to the same real-world event despite differences of their participants.

For event sequencing, there are few supervised methods on script-like relation classification due to the lack of data. To the best of our knowledge, the only work in this direction is by \newcite{Araki2013}. This work focuses on the other type of relations in the event sequencing task: \textbf{Subevent} relations. There is also a rich literature on unsupervised script induction \cite{Chambers2008,Cheung2013,Rudinger2015LM,Pichotta2016acl,Ferraro2016} that extracts scripts as a type of common-sense knowledge from raw documents. The focus of this work is to make use of massive collections of text documents to mine event co-occurrence patterns. In contrast, our work focuses on parsing the detailed relations between event mentions in each document.

Another line of work closely related to event sequencing is to detect other temporal relations between events. Recent computational approaches for temporal detection are mainly conducted on the TimeBank corpus \cite{Pustejovsky2002}. There have been several studies on building automatic temporal reasoning systems \cite{Uzzaman2010,Do2012,Chambers2014}. In comparison, the Event Sequencing task is motivated by the Script theory, which places more emphasis on common-sense knowledge about event chronology.

%% file: 3model.tex
\section{Model}
\subsection{Graph-Based Decoding Model}
\label{sec:graph_model}
In the Latent Antecedent Tree (LAT) model popularly used for entity coreference decoding~\cite{Fernandes2012,Bjorkelund2014}, 
each node represents an event mention and each arc a coreference relation, and new mentions are connected to some past mention considered most similar.  Thus the LAT model represents the decoding structure as a tree.  This can represent any coreference cluster, because coreference relations are by definition equivalence relations\footnote{An equivalence relation is reflexive, symmetric and transitive.}. 

In contrast, tree structures cannot always fully cover an Event Sequence relation graph, because 1) the After links are directed, not symmetric, and 2) multiple event nodes can link to one node, resulting in multiple parents. 

\begin{figure}
\centering
\includegraphics[width=0.45\textwidth]{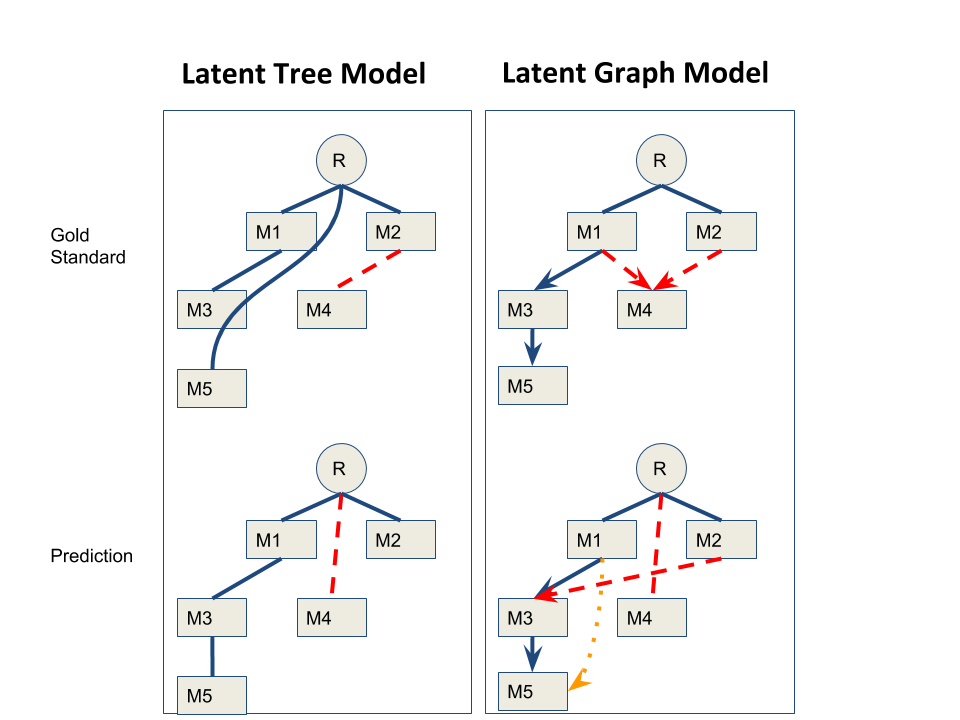}
\caption{\label{fig:model_compare} Latent Tree Model (left): tree structure formed by undirected links. Latent Graph Model (right): a DAG form by directed links. Dashed red links highlight the discrepancy between prediction and gold standard. The dotted yellow link (bottom right) can be inferred from other links.}
\end{figure}

To solve this problem, we extend the LAT model and propose its graph version, namely the Latent Antecedent Graph (LAG) model.
Figure \ref{fig:model_compare} contrast LAT and LAG with decoding examples. The left box shows two example decoded trees in LAT, where each node has one single parent. The right box shows two example decoded trees in LAG, where each node can be linked to multiple parents.

Formally, we define the series of (pre-extracted) event mentions of the document as $M=\{m_0, m_1, ..., m_n\}$, following their discourse order. $m_0$ is an artificial root node preceding all mentions. 
For each mention $m_j$, let $A_j$ be the set of its potential antecedents: $A_j = \{m_0, m_1, ..., m_{j-1}\}$. Let $\mathcal{A}$ denotes the set of antecedents for all the mentions in the sequence $\{A_0, A_1, ..., A_n\}$. 
The two tasks in question can be considered as finding the appropriate antecedent(s) from $\mathcal{A}$. Similarly, we define the gold antecedent set $\tilde{\mathcal{A}} = \{\tilde{A_0}, \tilde{A_1}, ..., \tilde{A_n}\}$, where $\tilde{A_i}$ represent the set of antecedents of $m_i$ allowed by the gold standard. In the coreference task, $\tilde{A_i}$ contains all antecedents that are coreferent with $m_i$. In the sequencing task, $\tilde{A_i}$ contains all antecedents that have an $After$ relation to $m_i$.

We can now describe the decoding process. We represent each arc as $\langle m_i, m_j, r \rangle (i < j)$, where $r$ is the relation name. The relation direction can be specified in the relation name $r$ (e.g. $r$ can be \textit{after.forward} or \textit{after.backward}). Further, an arc from the root node $m_0$ to node $m_j$ represents that $m_j$ does not have any antecedent. The score of the arc is the dot product between the weight parameter $\vec{w}$ and a feature vector $\Phi(\langle m_i, m_j, r\rangle)$, where $\Phi$ is an arc-wise feature function. The decoded graph $z$ can be determined by a set of binary variables $\vec{z}$, where $\vec{z}_{ijr} = 1$ if there is an arc $\langle m_i, m_j, r \rangle$ or 0 otherwise. The final score of $z$ is the sum of scores of all arcs:
\vspace{-0.1in}
\begin{align}
score(z) = \sum_{i,j, r} \vec{z}_{ijr} \vec{w} \cdot \Phi (\langle m_i, m_j, r\rangle)
\end{align}
The decoding step is to find the output $\hat{z}$ that maximizes the scoring function:
\vspace{-0.1in}
\begin{align}
\label{for:decoding}
\hat{z} = \arg\max_{z \in \mathcal{Z}(\mathcal{A})} score (z) 
\end{align}
where $\mathcal{Z}(\mathcal{A})$ denotes all possible decoding structures
given the antecedent sets $\mathcal{A}$. It is useful to note that the decoding step can be applied in the same way to the gold antecedent set $\tilde{\mathcal{A}}$.

Algorithm \ref{alg:pa_training} shows the Passive-Aggressive training algorithm \cite{Crammer2006} used in our decoding framework. Line \ref{line:decoding_gold} decodes the maximum scored structure from all possible gold standard structures using the current parameters $\vec{w}$. Intuitively, this step tries to find \textbf{the ``easiest'' correct graph} --- the correct graph with the highest score --- for the current model. Several important components remain unspecified in algorithm \ref{alg:pa_training}: (1) the decoding step (line \ref{line:decoding_sys}, \ref{line:decoding_gold}); (2) the match criteria: whether to consider the system decoding structure as correct (line \ref{line:match}); (3) feature delta: computation of feature difference (line \ref{line:delta}); (4) loss computation (line \ref{line:loss}). We detail the actual implementation of these steps in \S\ref{sec:training_details}.

\SetAlgoSkip{}
\begin{algorithm}
\caption{PA algorithm for training}
\label{alg:pa_training}
\textbf{Input:} Training data D, number of iterations T \\
\textbf{Output:} Weight vector $\vec{w}$ \\
$\vec{w} = \vec{0}$; \\
$\langle \mathcal{A}, \tilde{\mathcal{A}} \rangle \in D$;\\
\For {$t \gets 1..T$}{
     	$\hat{z} = \arg\max_{\mathcal{Z}(\mathcal{A})} score(z)$\; \label{line:decoding_sys}
    	\If {$\neg Match(\hat{z}, \tilde{\mathcal{A}})$}{\label{line:match}
          $\tilde{z} = \arg\max_{\mathcal{Z}(\tilde{\mathcal{A}})} score(z)$\; \label{line:decoding_gold}
          $\Delta = FeatureDelta(\tilde{z}, \hat{z})$\;\label{line:delta}
          $\tau = \frac{loss(\tilde{z}, \hat{z})}{||\Delta||^2}$\;\label{line:loss}
          $w = w + \tau\Delta$\;
        }
}
\Return w\;
\end{algorithm}

\subsubsection{Minimum Decoding Structure}
\label{sec:minimum_decoding}
Similar to the LAT model, there may be many decoding structures representing the same configuration. In LAT, since there is exactly one link per node, the number of links in different decoding structures is the same, hence comparable. In LAG, however, one node is allowed to link to multiple antecedents, creating a potential problem for decoding.  For example, consider the sequence $m_1 \xrightarrow[]{\text{after}} m_2 \xrightarrow[]{\text{after}} m_3$, both of the following structures are correct:

\begin{enumerate}
\item $\langle m_1, m_2, after\rangle$, $\langle m_2, m_3, after\rangle$
\item $\langle m_1, m_2, after\rangle$, $\langle m_2, m_3, after\rangle$, $\langle m_1, m_3, after\rangle$
\end{enumerate}

However, the last relation in the second decoding structure can actually be inferred via transitivity. We do not intend to spend the modeling power on such cases. We empirically avoid such redundant cases by using the \textbf{transitive reduction graph} for each structure. For a directed acyclic graph, a transitive reduction graph contains the fewest possible edges that have the same reachability relation as the original graph. In the example above, structure 1 is a transitive reduction graph for structure 2. We call the decoding structures that corresponding to the reduction graphs as \textit{minimum decoding structures}. For LAG, we further restrict $\mathcal{Z}(\mathcal{A})$ to contain only minimum decoding structures.

\subsubsection{Training Details in Latent Antecedent Graph}
\label{sec:training_details}
In this section, we describe the decoding details for LAG. Note that if we enforce a single antecedent for each node (as in our coreference model), it falls back to the LAT model \cite{Bjorkelund2014}.

\textbf{Decoding:} 
We use a greedy \textbf{best-first decoder} \cite{Ng2002a}, which makes a left-to-right pass over the mentions. The decoding step is the same for line \ref{line:decoding_sys} and \ref{line:decoding_gold}. The only difference is that we will use gold antecedent set ($\tilde{\mathcal{A}}$) at line \ref{line:decoding_gold}. For each node $m_j$, we keep all links that score higher than the root link $\langle 0, m_j, r\rangle$.

\textbf{Cycle and Structure Check:} Incremental decoding a DAG may introduce cycles to the graph, or violate the minimum decoding structure criterion. To solve this, we maintain a set $R(m_i)$ that is reachable from $m_i$ during the decoding process. We reject a new link ($\langle m_j, m_i \rangle$ if $m_j \in R(m_i)$) to avoid cycles. We also reject a redundant link ($\langle m_i, m_j \rangle$ if $m_j \in R(m_i)$) to keep a minimum decoding structure. Our current implementation is greedy, we leave investigations of search or global inference based algorithms to future work. 

\textbf{Selecting the Latent Event Mention Graph:} Note that sequence relations are on the event level. Given a unique event graph, it may still correspond to multiple mention graphs. In our implementation, we use a minimum set of event mentions to represent the full event graph by taking one single mention from each event. Following the ``easiest'' intuition, we select the single mention that will result in the highest score given the current feature weight $w$.

\textbf{Match Criteria:} We consider two graphs to match when their inferred graphs are the same. The inferred graph is defined by taking the transitive closure of the graph and propagate the links through the coreference relations. For example, in Figure \ref{fig:example}, the mention \predicate{fired} will be linked to two \predicate{killed} mentions after propagation. 

\textbf{Feature Delta:} In structural perceptron training ~\cite{Collins2002}, the weights are updated directly by the feature delta. For all the features $\tilde{f}$ of the gold standard graph $\tilde{z}$ and features $\hat{f}$ of a decoded graph $\hat{z}$, the feature delta is simply: $\Delta = \tilde{f} - \hat{f}$. However, a decoded graph may contain links that are not directly presented but inferable from the gold standard graph. For example, in Figure \ref{fig:model_compare}, the prediction graph has a link from  $M5$ to $M1$ (the orange arc), which is absent but inferable from the gold standard tree. If we keep these links when computing $\Delta$, the model does not converge well. We thus remove the features on the inferable links from $\hat{f}$ when computing $\Delta$.

\textbf{Loss:} We define the loss to be the number of different edges in two graphs.  Following \newcite{Bjorkelund2014}, we further penalize erroneous root attachment: an incorrect link to the root $m_0$ adds the loss by 2. For example, in Figure \ref{fig:model_compare} the prediction graph (bottom right) incorrectly links $m_4$ to Root and misses a link to $m_3$, which cause a total loss of 3.
In addition, to be consistent with the feature delta computation, we do not compute loss for predicted links that are inferable from the gold standard.

\subsection{Features}
\subsubsection{Event Coreference Features}
For event coreference, we design a simple feature set to capture syntactic and semantic similarity of arcs. The main features are summarized in Table~\ref{tab:coref_features}. In the TAC KBP 2015 coreference task setting, the event mentions are annotated with two attributes. There are 38 event types and subtype pairs (e.g., {\em Busness.Merge-Org, Conflict.Attack}). There also 3 realis type: events that actually occurred are marked as \textit{Actual}; events that are not specific are marked as \textit{Generic}; other events such as future events are marked as \textit{Other}. For these two attributes, we use the gold annotations in our feature sets.

\begin{table*}[t]
\centering
{
\begin{tabular}{l | l}\hline
Head & Headword token and lemma pair, and whether they are the same. \\\hline
Type & The pair of event types, and whether they are the same.\\\hline
Realis & The pair of realis types and whether they are the same. \\\hline
POS & POS pair of the two mentions and whether they are the same. \\\hline
Exact Match & Whether the 5-word windows of the two mentions matches exactly. \\\hline
Distance & Sentence distance between the two mentions. \\\hline
Frame & Frame name pair of the two mentions and whether they are the same. \\\hline
Syntactic & Whether a mention is the syntactic ancestor of another. \\\hline
\end{tabular}
}
\caption{\label{tab:coref_features} Coreference Features. Parsing is done using Stanford CoreNLP \protect\cite{Manning2014}; frame names are produced by Semafor \protect\cite{Das2011}.}
\end{table*}

\subsubsection{Event Sequencing Features}
\label{sec:features_sequencing}
An event sequencing system needs to determine whether the events are in the same script and order them. We design separate feature sets to capture these aspects: the Script Compatibility set considers whether  mentions should belong to the same script; the Event Ordering set determines the relative ordering of the mentions. Our final features are the cross products of features from the following 3 sets.

\begin{enumerate}
\item \textbf{Surface-Based Script Compatibility}: these features capture whether two mentions are script compatible based on the surface information, including:
\begin{itemize}
\item Mention headword pair.
\item Event type pair.
\item Whether two event mentions appear in the same cluster in Chambers's event schema database \cite{Chambers2010}.
\item Whether the two event mentions share arguments, and the semantic frame name of the shared argument (produced by the Semafor parser~\cite{Das2011}).
\end{itemize}

\item \textbf{Discourse-Based Script Compatibility}: these features capture whether two event mentions are related given the discourse context.
\begin{itemize}
\item Dependency path between the two mentions.
\item Function words (words other than Noun, Verb, Adjective and Adverb) in between the two mentions.
\item The types of other event mentions between the two mentions.
\item The sentence distance of two event mentions.
\item Whether there are temporal expressions (AGM-TMP slot from a semantic parser \cite{Tratz2011}) in the sentences of the two mentions.
\end{itemize}

\item \textbf{Event Ordering}: this feature set tries to capture the ordering of events. We use the discourse ordering of two mentions (forward: the antecedent is the parent; backward: the antecedent is the child), and temporal ordering produced by Caevo \cite{Chambers2014}.
\end{enumerate}

Taking the \textit{after} arc from \predicate{fired} to \predicate{killed} in Figure~\ref{fig:example} as an example, a feature after the cross product is: Event type pair is {\em Conflict.Attack} and {\em Life.Die}, discourse ordering is {\em backward}, and sentence distance is 0. 

%% file: 4experiment.tex
\section{Experiments}
\subsection{Dataset}
We conduct experiments on the dataset released in Text Analysis Coreference (TAC-KBP) 2017 Event Sequencing task (released by LDC under the catalog name LDC2016E130). This dataset contains rich event relation annotations, with event mentions and coreference annotated in TAC-KBP 2015, and additional annotations on Event Sequencing\footnote{http://cairo.lti.cs.cmu.edu/kbp/2016/after/}. There are 158 documents in the training set and 202 in the test set, selected from general news articles and forum discussion threads. The event mentions are annotated with 38 type-subtype and 3 realis status (Actual, Generic, Other). Event Hopper, After, and Subevent links are annotated between event mentions. For all experiments, we develop our system and conduct  ablation studies using 5-fold cross-validation on the training set, and report performance on the test set.

\subsection{Baselines and Benchmarks}
\textbf{Coreference:} we compare our event coreference system against the top performing systems from TAC-KBP 2015 (LCC, UI-CCG, and LTI). In addition, we also compare the results against two official baselines~\cite{Mitamura2016}: the Singleton baseline that put each event mention in its own cluster and the Match baseline that creates clusters based on mention type and realis status match. \\
\textbf{Sequencing:} This work is an initial attempt to this problem, so there is currently no comparable prior work on the same task. We instead compare with a baseline using event temporal ordering systems. We use a state-of-the-art temporal system named Caevo \cite{Chambers2014}. To make a fair comparison, we feed the gold standard event mentions to the system along with mentions predicted by Caevo\footnote{We keep the mentions predicted by Caevo because its inference may be affected by these mentions.}. However, since the script-style After links are only connected between mentions in the same script, directly using the output of Caevo produces very low precision. Instead, we run a stronger baseline: we take the gold standard script clusters and then only ask Caevo to predict links within these clusters (Oracle Cluster + Temporal).

\subsection{Evaluation Metrics}
\label{sec:metric}
\textbf{Evaluating Event Coreference:} We evaluate our results using the official scorer provided by TAC-KBP, which uses 4 coreference metrics: \textit{BLANC} \mbox{\cite{Recasens2011}}, \textit{MUC} \cite{Chinchor1992}, \textit{$B^3$} \cite{Bagga1998} and \textit{CEAF-E} \cite{Luo2005}. Following the TAC KBP task, systems are ranked using the average of these 4 metrics. 

\noindent\textbf{Evaluating Event Sequencing:} The TAC KBP scorer evaluates event sequencing using the metric of the TempEval task \cite{UzZaman2012,UzZaman2013}. The TempEval metric calculates special precision and recall values based on the closure and reduction graphs:
\begin{align*}
\label{for:eval}
Precision = \frac{|Response^- \cap Reference^+|}{|Response^-|}  \quad
Recall = \frac{|Reference^- \cap Response^+|}{|Reference^-|}
\end{align*}
where $Response$ represents the After link graph from the system response and $Reference$ represents the After link graph from the gold standard. $G^+$ represents the graph closure for graph $G$ and $G^-$ represents the graph reduction for graph $G$. 
As preprocessing, relations are automatically propagated through coreference clusters (currently using gold standard clusters). The final score is the standard F-score: geometric mean of the precision and recall values.

\subsection{Evaluation Results for Event Coreference}
\label{sec:coref_results}
The test performance on Event Coreference is summarized in Table \ref{tab:coreference_test}. Comparing to the top 3 coreference systems in TAC-KBP 2015, we outperform the best system by about 2 points absolute F-score on average. Our system is also competitive on individual metrics. Our model performs the best based on $B^3$ and CEAF-E, and is comparable to the top performing systems on  MUC and BLANC. 

Note that while the \texttt{Matching} baseline only links event mentions based on event type and realis status, it is very competitive and performs close to the top systems. This is not surprising since these two attributes are based on the gold standard. To take a closer look, we conduct an ablation study by removing the simple match features one by one. The results are summarized in Table~\ref{tab:coref_ablation}. We observe that some features produce mixed results on different metrics: they provide improvements on some metrics but not all. This is partially caused by the different characteristics of different metrics. On the other hand, these features (parsing and frames) are automatically predicted, which make them less stable. Furthermore, the Frame features contain duplicate information to event types, which makes it less useful in this setting.

Besides the presented features, we have also designed features using event argument. However, we do not report the results since the argument features decrease the performance on all metrics. 

\begin{table}[t]
  \centering
  \begin{tabular}{l | r | r | r | r | r }
  \hline
   & $B^3$ & CEAF-E & MUC & BLANC & AVG. \\\hline
     \texttt{Singleton} & 78.10 & 68.98 & 0.00 & 48.88 & 52.01 \\
  \texttt{Matching} & 78.40 & 65.82 & \textbf{69.83} & 76.29 & 71.94 \\ \hline
  LCC & 82.85 & 74.66 & 68.50 & \textbf{77.61} & 75.69 \\
  UI-CCG & 83.75 & 75.81 & 63.78 & 73.99 & 74.28 \\
  LTI & 82.27 & 75.15 & 60.93 & 71.57 & 72.60 \\\hline
  This work & \textbf{85.59} & \textbf{79.65} & 67.81 & 77.37 & \textbf{77.61} \\
  \hline
  \end{tabular}
  \caption{\label{tab:coreference_test}Test Results for Event Coreference with the \texttt{Singleton} and \texttt{Matching} baselines.}
\vspace{1em}
  \begin{tabular}{l | r | r | r | r | r }
  \hline
   & $B^3$ & CEAF-E & MUC & BLANC & AVG. \\\hline
  ALL & 81.97 & 74.80 & 76.33 & 76.07 & 77.29 \\\hline
  -Distance & 81.92 & 74.48 & 76.02 & 77.55 & 77.50 \\\hline
  -Frame & 82.14 & 75.01 & 76.28 & 77.74 & 77.79  \\\hline
  -Syntactic & 81.87 & 74.89 & 75.79 & 76.22 & 77.19 \\\hline
  \end{tabular}
  \caption{\label{tab:coref_ablation} Ablation study for Event Coreference.}
\end{table}

\subsection{Evaluation Results for Event Sequencing}
\label{sec:seq_results}
The evaluation results on Event Sequencing is summarized in Table \ref{tab:seq_test}. Because the baseline system has access to the oracle script clusters, it produces high precision. However, the low recall value shows that it fails to produce enough After links. Our analysis shows that a lot of After relations are not indicated by clear temporal clues, but can only be solved with script knowledge. In Example \ref{exp:baseline}, the baseline system is able to identify ``fled'' is after ``ousted'' from explicit marker ``after''. However, it fails to identify that ``extradited'' is after ``arrested'', which requires knowledge about prototypical event sequences.

\begin{example}
\label{exp:baseline}
Eight months after the [{\scriptsize \color{red} transport} fled] Ivory Coast when Gbagbo, the former president, was [{\scriptsize \color{red} End.Position} ousted] by the French military. Blé Goudé was subsequently [{\scriptsize \color{red} Jail} arrested] in Ghana and [{\scriptsize \color{red} transport} extradited]
Megrahi,[{\scriptsize \color{red} Jail} jailed] for [{\scriptsize \color{red} Attack} killing] 270 people in 1988. \footnote{The small red text indicates the event type for each mention.}
\end{example}

In our error analysis, we noticed that our system produces a large number of relations due to coreference propagation. One single wrong prediction can cause the error to propagate. 

Besides memorizing the mention pairs, our model also tries to capture script compatibility through discourse signals. To further understand how much these signals help, we conduct an ablation study of the features in the discoursed based compatibility features (see \S\ref{sec:features_sequencing}). Similarly, we remove each feature group from the full feature set one by one and observe the performance change.

The results are reported in Table \ref{tab:ablation}. While most of the features only affect the performance by less than 1 absolute F1 score, the feature sets after removing \textit{mention} or \textit{sentences} show a significant drop in both precision and recall. This shows that discourse proximity is the most significant ones among these features. In addition, the \textit{mention} feature set captures the following \textit{explain away} intuition: the event mentions A and B are less likely to be related if there are similar mentions in between. One such example can be seen in Figure \ref{fig:example}, the event mention \predicate{fired} is more likely to relate to the closest \predicate{killed}, instead of the other \predicate{killed} in the first paragraph.

In addition, our performance on the development set is higher than the test set. Further analysis reveals two causes: 1) the coreference propagation step causes the scores to be very unstable,  2) our model only learns limited common sense ordering based on lexical pairs, which overfit to the small training corpus. Since the annotation is difficult to scale, it is important to use methods to harvest script common sense knowledge automatically, as in the script induction work~\cite{Chambers2008}.

\begin{table}[t]
\centering
{
\begin{tabular}{l | l | l | l }
\hline
& Prec. & Recall & F-Score    \\ \hline
Oracle Cluster+Temporal & \textbf{46.21} & 8.72 & 14.68 \\\hline
Our Model & 18.28 & \textbf{16.91} & \textbf{17.57} \\ \hline
\end{tabular}
}
\caption{\label{tab:seq_test}Test Results for event sequencing. The Oracle Cluster+Temporal system is using Caevo's result on the Oracle Clusters.}
\end{table}

\begin{table}[t]
\centering
{
\begin{tabular}{l | l | l | l | l} 
\hline
 & Prec. & Recall & F-Score & $\Delta$\\\hline
Full  & 37.92	&	36.79	&	36.36 \\\hline
- Mention Type & 32.78	&	29.81	&	30.07	&	6.29 \\\hline
- Sentence & 33.90	&	30.75	&	31.00	&	5.36 \\\hline
- Temporal & 37.21	&	36.53	&	35.81	&	0.55 \\\hline
- Dependency & 38.18	&	36.44	&	36.23	&	0.13 \\\hline
- Function words & 38.08	&	36.51	&	36.18	&	0.18\\\hline
\end{tabular}
}
\caption{\label{tab:ablation}Ablation Study for Event Sequencing.}
\end{table}

%% file: 5discussion.tex
\section{Discussion}
\label{sec:discussion}
\subsection{Event Coreference Challenges}
Although we have achieved good performance on event coreference, upon closer investigation we found that most of the coreference decisions are still made based on simple word/lemma matching (note that the type and realis baseline is as high as 0.72 F1 score). The system exploits little semantic information to resolve difficult event coreference problems. A major challenge is that our system is not capable of utilizing event arguments: in fact, \newcite{Hasler2009} found that only around 20\% of the arguments in the same event slot are actually coreferent for coreferential event pairs in the ACE 2005 corpus. Furthermore, the TAC-KBP corpus uses a relaxed participant identity requirement for event coreference, which makes argument-based matching more difficult.

\subsection{Event Sequencing Challenges}
Our event sequencing performance is still low despite the introduction of many features. This task is inherently difficult because it requires a system to solve both the script clustering and event ordering tasks. The former task requires both common-sense knowledge and discourse reasoning. Reasoning is more important for long-term links since there are no explicit clues like prepositions and dependencies to be exploited. The ablation study shows that discourse features like sentence distance are more effective, which indicates that our model mainly relies on surface clues and has limited reasoning power.

Furthermore, we observe a strong locality property of After links by skimming the training data: most After link relations are found in a small local region. Since reasoning and coreference based propagation will accumulate local decisions, a system must be accurate on them.

\subsubsection{The Ambiguous Boundary of a Script}
Besides the above-mentioned challenges, a more fundamental problem is to define the boundary of scripts. Since the definition of scripts is only prototypical event sequences, the boundaries between them are not clear. 
In Example \ref{exp:baseline}, the event \predicate{jailed} is considered to belong to a ``Judicial Process'' script and \predicate{killing} is considered to belong to an``Attack'' script\footnote{Script names are taken from the annotation guideline: \url{http://cairo.lti.cs.cmu.edu/kbp/2016/after/annotation}}. No link is annotated between these two mentions since they are considered to belong to different clusters, even though the ``jailed'' event is to punish the ``killing''. Therefore essentially, the current Event Sequencing task simply requires the system to fit these human defined boundaries. In principle, the ``Judicial Process'' script and the ``Attack'' script can form a larger script structure, on a higher hierarchical level.


While it is possible to manually define scripts and what kind of events they may contain specifically in a controlled domain, it is difficult to generalize the relations. Most previous work on script induction \cite{Chambers2008,Cheung2013,Rudinger2015LM,Pichotta2016acl,Ferraro2016} treats scripts as statistical models where probabilities can be assigned, thereby avoiding the boundary problem.
While the script boundaries may be application dependent, a possible solution may rely on the ``Goals'' in Schank's script theory. The Goal of a script is the final state expected (by the script protagonist) from the sequence of events. Goal oriented scripts may be able to help us explain whether \predicate{killing} and \predicate{jailed} should be separate: if we take the``killer'' as the protagonist, the goal of ``kill'' is achieved at the point of the victim dying. We leave the investigation on proper theoretical justification to future work.

%% file: 6conclusion.tex
\section{Conclusion}
In this paper, we presented a unified graph framework to conduct event coreference and sequencing. We have achieved state-of-the-art results on event coreference and report the first attempt at event sequencing. While we only studied two types of relations, we believe the method can be adopted in broader contexts. 
In the future, we plan to build a joint model to allow the tasks to mutually improve each other.

In general, analyzing event structure can bring new aspects of knowledge from text. For instance, Event Coreference systems can help group scattered information together. Understanding Event Sequencing can help clarify the discourse structure, which can be useful in other NLP applications, such as solving entity coreference problems \cite{Peng2015}. However, in our investigation, we find that the linguistic theory and definitions for events are not adequate for the computational setting. For example, proper theoretical justification is needed to define event coreference, which should explain the problems, such as argument mismatches. In addition, we also need a theoretical basis for script boundaries. In the future, we will devote our effort to understanding the theoretical and computational aspects of events relations, and utilizing them for other NLP tasks. 
